\pdfoutput=1

\documentclass[11pt]{article}

\usepackage{acl}

\usepackage{times}
\usepackage{latexsym}
\usepackage{graphicx}
\usepackage[normalem]{ulem}

\definecolor{lightgray}{gray}{0.95}

\usepackage{microtype}
\usepackage{graphicx}
\usepackage{subfigure}
\usepackage[subfigure]{tocloft}
\usepackage{booktabs} 
\usepackage{amsmath}
\usepackage{amssymb}
\usepackage{amsfonts}
\usepackage{multirow}
\usepackage{verbatim}
\usepackage{caption}
\usepackage{longtable}
\usepackage{float}

\usepackage{enumitem}
\usepackage{tablefootnote}
\usepackage{xcolor}
\usepackage{xspace}
\usepackage{textcomp}
\usepackage{makecell}
\usepackage{multirow}
\usepackage{lscape} 
\usepackage{siunitx}

\usepackage{amssymb}
\usepackage{pifont}
\usepackage{scrextend}
\usepackage{todonotes}

\newcommand{\palmname}[0]{{PaLM}\xspace}
\newcommand{\stmoename}[0]{{ST-MoE}\xspace}
\newcommand{\lamdaname}[0]{{LaMDA}\xspace}

\usepackage[T1]{fontenc}

\usepackage[utf8]{inputenc}

\usepackage{microtype}
\usepackage{multirow}
\usepackage{amsfonts,amsmath,bm}
\title{Benchmarking Large Language Model Capabilities for Conditional Generation}

\author{
  Joshua Maynez \\  Google DeepMind \\ {\small\tt joshuahm@google.com} \And
  Priyanka Agrawal \\ Google DeepMind \\ {\small\tt priyankagr@google.com}  \AND
  Sebastian Gehrmann \\ Google Research \\ {\small\tt gehrmann@google.com}
}

\begin{document}
\maketitle

\begin{abstract}
Pre-trained large language models (PLMs) underlie most new developments in natural language processing. They have shifted the field from application-specific model pipelines to a single model that is adapted to a wide range of tasks. Autoregressive PLMs like GPT-3 or PaLM, alongside techniques like few-shot learning, have additionally shifted the output modality to generation instead of classification or regression. 
Despite their ubiquitous use, the generation quality of language models is rarely evaluated when these models are introduced. Additionally, it is unclear how existing generation tasks–--while they can be used to compare systems at a high level–--relate to the real world use cases for which people have been adopting them. In this work, we discuss how to adapt existing application-specific generation benchmarks to PLMs and provide an in-depth, empirical study of the limitations and capabilities of PLMs in natural language generation tasks along dimensions such as scale, architecture, input and output language. Our results show that PLMs differ in their applicability to different data regimes and their generalization to multiple languages and inform which PLMs to use for a given generation task setup. We share best practices to be taken into consideration when benchmarking generation capabilities during the development of upcoming PLMs. 
\end{abstract}

\section{Introduction}
\label{sec:introduction}

Natural language generation tasks require generating understandable text given textual or non-linguistic information as input, such as documents, tables, or other structured forms. These texts seek to achieve a communicative goal (e.g., summarize a document). 
The standard approach to tackle these problems over the last years has been to start with a pretrained encoder-decoder model like T5~\citep{Raffel:ea:2019} or BART~\citep{Lewis2020BARTDS} and finetune it on a corpus that captures the downstream task. 
The recent much larger pretrained language models use a decoder-only architecture and upended this paradigm.
These models enabled few-shot or in-context learning approaches in which a model is presented with one or more examples and tasked to continue generating without any finetuning. We refer to both kinds of pretrained models as PLMs.

Due to the lack of grounding in the specific task setup, few-shot learning in generation settings leads to a model approaching the communicative goal from very different angles. These diverse range of outputs make the typical reference-based automatic evaluation strategies largely incompatible.
While human evaluation can be used to overcome this shortcoming, it is infeasible to monitor the performance of an actively training model this way or to re-run all evaluations every time a new model is introduced. 
This leads to the question how one should reliably monitor generation capabilities, a question that is only growing in importance as more tasks are approached by casting them into generation setups. 

In this work, we evaluate 8 models in few-shot and finetuning settings on 27 generation tasks covering 14 languages via automatic evaluation, presenting the first large-scale benchmark of PLMs in conditional NLG settings.
We discuss design choices and challenges to ensure a fair comparison between the different systems, including  suitable methods, tasks, and metrics. 
Based on our empirical results, we derive recommendations that could be used for future benchmarks during the development of PLMs.
To combat the need for repeating computationally expensive explorations, we investigate how many evaluation examples are necessary to identify differences between models and find that, in many cases, fewer than 500 examples are sufficient, which opens the path for future evaluation-only task developments. 



\section{Background and Related Work}
\label{sec:related-work}

The shift from specialized pipelines toward pretrained language models has led to significant changes in how models are evaluated. We now focus more on questions such as ``\textit{how good are the learned representations?}'' instead of user-facing measures of utility. The changes manifested in leaderboards and standard benchmarks that aim to characterize a wide range of model capabilities~\citep{ethayarajh-jurafsky-2020-utility}. 

An additional recent shift is that from finetuning toward few-shot learning. Models like T5~\cite{Raffel:ea:2019}, BART~\cite{Lewis2020BARTDS}, and mT5~\cite{xue-etal-2021-mt5} were finetuned on supervised datasets covering tasks including translation and summarization, and their outputs are compared to ``ground truth'' outputs via widely used metrics like ROUGE~\citep{lin-2004-rouge} which provide a noisy indication of the ``quality'' of the output and which can be used to determine whether a model is better than others.\footnote{For an in-depth review of the usefulness of automatic metrics, we refer to \citet{Gehrmann2022RepairingTC} and point to Section~\ref{sec:results} for a discussion of the application of metrics to benchmarks.}
In contrast, large PLMs with autoregressive language modeling pretraining objectives are more capable to produce results without explicit finetuning and are thus typically evaluated via few-shot and in-context approaches, where the model is given the task description and exemplars showing how the task should be completed.
GPT-3 \cite{Brown2020LanguageMA} and models that followed such as GLaM \cite{Du2022GLaMES}, Gopher \cite{Rae2021ScalingLM}, and LaMDA \cite{Thoppilan2022LaMDALM}, have achieved few-shot state-of-the-art results on a large number of tasks at their time of publication. 
However, few-shot approaches work best for tasks with a clear answer such as classification or span-based question-answering.\footnote{We refer to the two task types as NLU and NLG tasks but note that this distinction becomes fuzzy with autoregressive models since technically all answers are ``generated''.}


Generation metrics penalize systems when their writing style differs from how the references are written~\cite{mathur-etal-2020-results,freitag-etal-2020-human,DBLP:conf/nips/MilleDMPGKMG21}. Without finetuning, there is no guarantee that PLMs produce outputs that look like the ground truth, both in style and content. 
Recent work found that these differences leads to sharp differences in how humans and automatic metrics rate the generation quality~\citep{DBLP:journals/corr/abs-2209-12356}. 
Due to this uncertainty, most evaluations of new PLMs are limited to NLU benchmarks such as SuperGLUE \cite{Wang2019SuperGLUEAS}. For example, LaMDA \cite{Thoppilan2022LaMDALM} did not evaluate on NLG tasks, GLaM \cite{Du2022GLaMES} limited its generation evaluation to short span question answering tasks, and GPT-3 \cite{Brown2020LanguageMA} evaluated only on machine translation. A first autoregressive PLM with broad NLG evaluation, PaLM \cite{Chowdhery2022PaLMSL}, benchmarked summarization and data-to-text tasks in multiple languages.

The recent Holistic Evaluation of Language Models project \cite[HELM,][]{Liang2022HolisticEO} aims to standardize evaluation of language models. With the explicit goal to broaden the task and metric coverage, HELM has established an impressive few-shot benchmark for many natural language tasks. Corroborating the prior findings, they also conclude that human evaluation is necessary for NLG. 
This distinction means that the reference-based approach for generated text that the field has used since the advent of deep learning may no longer sufficient and that we need clear evaluation protocols that continue to allow us to answer broad questions about ``generation quality'' of a model.
Complementing this work, we take a deeper look at a wider set of NLG tasks and explore LLMs in finetuning and few-shot setups to identify whether reference-based automatic evaluation can still be used to produce system rankings.




\paragraph{Research Questions}
We aim to define a methodology that allows us to answer the question ``How good are learned representations of a model for generating natural language?'' via few-shot and finetuning approaches. To develop and apply this methodology we seek to answer the following three research questions:

\begin{itemize}[leftmargin=5mm,noitemsep]

    
    \item[\textbf{R1}] \textit{How do different model architectures compare in terms of automatic metrics?}
    
    We aim to identify patterns that emerge in evaluations and to uncover aspects inherent to the tasks, e.g. \textit{have metrics on specific tasks saturated?}, and to the models' architectural choices, e.g., are encoder-decoders better suited for particular task formulations? (Section~\ref{sec:results})
    
    \item[\textbf{R2}] \textit{What set of tasks, methods, and metrics is best suited for the monitoring of improvements in language generation capabilities?}
    
    Using the results of R1, we aim to select a subset of tasks, methods, and metrics that robustly produce reliable model rankings. (Section~\ref{sec:practices})

    \item[\textbf{R3}] \textit{What are the broader implications for how the quality of newly developed models should be monitored?}
    
    Robustly ranking systems is particularly important when monitoring a system during training and when comparing across many tasks. In line with the ``reality check'' theme track at ACL 2023, we discuss the implications of our findings on how evaluation results should be produced and interpreted.  (Section~\ref{sec:discussion})
     
\end{itemize}


\section{Method}
\label{sec:method}


\subsection{Data}

We select a combination of data-to-text and text-to-text datasets as different input modalities. 
The selected datasets capture different input and output lengths, domains, languages, and communicative goals. 
The text-to-text task with most available multilingual datasets is summarization which we pick for this paper.\footnote{Since benchmarks for machine translation are well-established~\citep[e.g.,][]{akhbardeh-etal-2021-findings} we exclude it from our scope. However, any methodological outcomes of our work can be applied to translation or similar tasks.}
We pick the following tasks:\footnote{All datasets were retrieved via the Generation Evaluation and Metrics benchmark~\citep{gehrmann2021gem,gehrmann2022gemv2}. We use these datasets for research purposes only in line with their intended use.} 

\begin{table}[t]
    \centering
    \resizebox{\columnwidth}{!}{%
    \begin{tabular}{ll|rr|rr}
    \toprule
    & & \multicolumn{2}{c}{Length} & \multicolumn{2}{c}{Size} \\
    \cmidrule(l{3pt}r{3pt}){3-4} \cmidrule(l{3pt}r{3pt}){5-6}
    Dataset & Languages & Input & Output & Training & Test \\ 
    \midrule
    E2E & en   & 146 & 135 & 35k & 4.7k     \\
    WebNLG & en,ru  & 169.5 & 157&  14k–35k & 1.1k-1.8k   \\
    ToTTo & en  & 357  & & 120k & 7.7k    \\
    Czech Rest. & cs   & 70 & 80 &  3.5k& 842   \\
    XSum & en & 1845 & 153  & 23k & 1.2k    \\
    WikiLingua & en,es,ru,tr,vi  & 1k–5k & 159–489 &  5k–3.8M & 900-29k    \\
    MLSum & es,de &  4152 & 147 & 220k–250k & 10k-13k   \\
    XL-Sum  &  {ar,bn,ja,id,sw,}  & 1k–10k & 137–614 & 1.3k–300k & 500-9k  \\
    & {ko,ru,te,th,tr,}& & & & \\
    & es,vi,hi & & & & \\
    \bottomrule
     \end{tabular}%
     }
    \caption{Details of the datasets evaluated in this paper: languages, lengths in number of tokens according to the mT5 tokenizer~\cite{xue-etal-2021-mt5}, and size of the training and test splits.}
    \label{tab:data-table}
\end{table} 

\begin{itemize}[noitemsep,leftmargin=*]
    \item \textbf{MLSum}~\citep{scialom2020mlsum} -- Summarize a news article in multiple sentences.
    \item \textbf{WikiLingua}~\citep{ladhak2020wikilingua} -- Generate section headers for step-by-step instructions from WikiHow.
    \item \textbf{XSum}~\citep{narayan-etal-2018-dont} -- Generate the first sentence of a news article.
    \item \textbf{Clean E2E NLG}~\citep{novikova-etal-2017-e2e,e2e_cleaned} -- Given a set of key-value attribute pairs, describe a restaurant in one or two sentences.
    \item \textbf{Czech Restaurant response generation}~\citep{Dusek2019NeuralGF} -- Given a dialog context and a dialog act representation, generate a one sentence long response.
    \item \textbf{WebNLG 2020}~\citep{gardent2017creating,castro-ferreira20:bilin-bi-direc-webnl-shared} -- Verbalize subject-predicate-object triples in one or more sentences.
    \item \textbf{ToTTo}~\citep{Parikh2020ToTToAC} -- Describe highlighted cells in a table in a single sentence. 
    \item \textbf{XL-Sum}~\citep{hasan-2021-xl} -- Summarize a news article, in the same language, in a single sentence.
\end{itemize}

\noindent Table \ref{tab:data-table} provides an overview of these datasets in terms of languages, the lengths of input and output and split sizes. For highly multilingual datasets, we evaluate on a subset of typologically diverse languages following the selection by \citet{clark-etal-2020-tydi}. To this selection, we add languages that appear  bothin WikiLingua and XL-Sum.

\subsection{Models}

Prior results for the benchmarked tasks primarily come from finetuning T5 \citep{raffel2020exploring}, mT5 \citep{xue-etal-2021-mt5}, or BART \citep{lewis-etal-2020-bart}, which are encoder-decoder models pretrained with an infilling objectives. 
These models are significantly smaller than newer models like GPT-3, with sizes ranging from 130M to 13B parameters. Encoder-decoder models trained for infilling often outperform larger decoder-only LMs in the finetuning setting~\citep{tay2022scaling}, while the latter work better for few-shot setting. There has also been recent work on reducing the computational cost of large models by $\sim$10x by using a mixture of experts~\citep{zohp_stmoe_2022}. 
It is important to compare these diverse set of models to understand how scale plays a role with the model’s architecture and its pretraining. We benchmark the following models:\footnote{Model names omitted for anonymity.} 

\begin{itemize}[noitemsep,leftmargin=*]
    \item \textbf{\palmname}
    \palmname is a pretrained decoder-only transformer-based model trained with standard left-to-right language modeling objective. It is pretrained on a range of multilingual corpora including Wikipedia, news, and code. In this work, we use two models scales: 8B parameters and 540B parameters.
    \item \textbf{GPT-3.5}~\cite{Ouyang2022TrainingLM} GPT-3.5 is a 175B parameter decoder-only transformer-model of the GPT-3 family \citep{Brown2020LanguageMA} but trained on a blend of text and code from before Q4 2021. This model, named code-davinci-002, was introduced as the base model for InstructGPT-3 \citep{Ouyang2022TrainingLM} without the supervision on human-written demonstrations and human-vetted model samples.\footnote{More details can be found at https://beta.openai.com/docs/model-index-for-researchers}
    \item \textbf{\stmoename}~\cite{zohp_stmoe_2022}
    \stmoename is a 269B sparse pretrained variant of a dense encoder-decoder transformer-based model.
    \item \textbf{\lamdaname}~\cite{Thoppilan2022LaMDALM} 
    \lamdaname (137B parameters) is a decoder-only transformer-based language model specialized for dialog applications. It is pretrained on dialog data as well as web text data followed by rank-based tuning. 
    \item \textbf{T5}~\cite{Raffel:ea:2019} T5-XXL (11B parameters) is a pretrained encoder-decoder transformer-based model trained on a span corruption objective with a novel unified text-to-text format. It is pretrained on Common Crawl data, mostly containing English-only documents.
    \item \textbf{mT5}~\cite{xue-etal-2021-mt5} mT5-XXL (11B parameters) is a multilingual variant of T5 that was pretrained on a multilingual corpus, mC4, covering 101 languages. 
    \item \textbf{LongT5}~\cite{Guo:ea:2022} LongT5 (3B parameters) a similar architecture as T5, where the encoder is extended to have global-local attention sparsity patterns to handle long inputs. 
\end{itemize}

\subsection{Few-shot evaluation methodology}

To evaluate the models for few-shot inference, we concatenate a task-specific prompt\footnote{For Summarization, this prompt was \textit{``Summarize the following article:''}, and for Data-to-Text it was \textit{``Verbalize:''}. This was translated to the appropriate language.} to the input and prepend an output prompt to the output. To handle the oftentimes very long inputs or outputs for tasks such as summarization, inputs were truncated to 2048 tokens and inference was done providing only one exemplar at a time, referred to as 1-shot. These simple prompts are analogous to those used in related work \citep{Chowdhery2022PaLMSL,scao2022bloom}. We do not tune the prompts or use more complex strategies to keep fair comparisons between multiple systems, as prompt selection can lead to overfitting.  
The exemplars are separated through double linebreaks, which are also used to truncate output predictions for evaluation. 
All few-shot exemplars are randomly sampled from the training corpus. From early experimentation, we found this particularly important since it avoids overfitting to exemplars that work well for one model but not another.

\subsection{Finetuning methodology}

To use the decoder-only architectures during finetuning, inputs and targets are concatenated.
The concatenated sequences are  truncated to 2048 tokens, the training context used during pretraining, with 512 tokens reserved for the target. Only summarization tasks required input truncation. We finetuned models with standard hyperparameters; refer to Appendix-\ref{sec:appendix_technical} for thorough details. The best model checkpoint for each dataset was selected by the best performing geometric mean of ROUGE-1, ROUGE-2 and ROUGE-L scores on the validation set. Decoding was done with beam-search with a beam size of 4 for encoder-decoder models, while inference in decoder-only PLMs (\lamdaname, \palmname, \stmoename) was performed using top-k sampling with $k$=10, due to issues with scaling beam search at the time of publication.

\subsection{Metrics}

Following the suggestions by \citet{Gehrmann2022RepairingTC}, we report a combination of lexical and learned metrics, starting with ROUGE-2 and ROUGE-L~\citep{lin-2004-rouge}.
Since the default ROUGE implementation uses English-specific tokenization, stemming and punctuation normalization, it is incompatible with other languages. \citet{hasan-2021-xl} extended ROUGE by integrating additional stemmers and tokenizers to cover up to the 45 languages. To support more languages, and avoid dependency on varying implementations, we use a SentencePiece tokenizer~\cite{kudo-richardson-2018-sentencepiece} which, provided a vocabulary distribution file, is self-contained and has sensible fall-backs to unexpected words. Specifically, we used mT5's SentencePiece vocabulary.

For the same reason, we also evaluate with ChrF~\citep{popovic-2015-chrf}, which is a character-level n-gram overlap metrics and thus independent from tokenizers. BLEURT \cite{Sellam2020BLEURTLR,pu-etal-2021-learning} is a multilingual model-based evaluation metric for generation designed to compute the similarity between a pair of sentences i.e. a reference and a candidate. It finetunes RemBERT~\citep{chung2021rethinking} on synthetic sentence pairs and gold ratings. In contrast to the lexical metrics, BLEURT is meant to capture the non-trivial semantic similarities between two texts.

For brevity, the main text of this section focuses on the F-measure of ROUGE-L for English and SentencePiece-ROUGE-L for all other languages while the remaining results are in Appendix~\ref{sec:appendix}. We additionally investigate the agreement between metrics in Section~\ref{sec:practices}.\footnote{For ROUGE, we used the python implementation at \url{https://github.com/google-research/google-research/tree/master/rouge} at commit \texttt{f935042} and whitespace-tokenized references and predictions before calling the library. For BLEURT, we used BLEURT-20 checkpoint from the library at \url{https://github.com/google-research/bleurt} and commit c6f2375.}

\section{Empirical Observations}
\label{sec:results}


\begin{table*}[h!]
\footnotesize{
    \centering
    \begin{tabular}{p{3cm}cccc|cccccc}
    \toprule
    & \multicolumn{4}{c}{One-shot} & \multicolumn{6}{c}{Finetuning} \\
    \cmidrule(l{3pt}r{3pt}){2-5} \cmidrule(l{3pt}r{3pt}){6-11}
    Task & \scriptsize{\makecell[c]{\palmname \\8B}} & \scriptsize{\makecell[c]{\palmname \\540B}} & \scriptsize{\makecell[c]{\lamdaname \\137B}} & \scriptsize{\makecell[c]{GPT-3.5 \\175B}}  & \scriptsize{\makecell[c]{\palmname \\8B}} & \scriptsize{\makecell[c]{\palmname \\540B}} & \scriptsize{\makecell[c]{\stmoename \\32B}} &  \scriptsize{\makecell[c]{T5 \\11B}} & \scriptsize{\makecell[c]{mT5 \\11B}} & \scriptsize{\makecell[c]{LongT5 \\3B}}   \\
    \midrule
     & \multicolumn{7}{c}{\textbf{Data-To-Text}} \\
    \midrule
    E2E (en) & 37.7 & \textbf{46.6} & 7.1 & \textbf{46.6} & 52.9 & 52.3 & 51.5 & 52.9 & 52.2 & \textbf{53.1}  \\
    WebNLG (en) & 45.3 & \textbf{54.7} & 8.4 & 54.6 & 56.8 & \textbf{58.0} & 56.4 & 50.8 & 47.7 & \textbf{58.0}  \\
    ToTTo (en) & 40.2 & \textbf{50.7} & 5.6 & \textbf{51.9} & 65.8 & \textbf{67.5} & 67.0 & 66.1 & 65.5 & 66.3  \\
    Czech Restaurant (cs) &  16.9 & 34.1 & 3.3 & \textbf{38.5} & \textbf{45.5} & \textbf{45.5} & 40.7 & 45.4 & 39.4 & 44.8 \\
    WebNLG (ru) & 16.8 & \textbf{33.7} & 4.5  & 33.3 & 40.9 & 40.5 & 28.2 & 41.2 & 41.1 & \textbf{41.6}  \\
    \midrule
     & \multicolumn{7}{c}{\textbf{English Generation}} \\
    \midrule
    XSum (en) & 19.9 & 28.6 & 10.0 & \textbf{34.0} & 31.4 & 36.5 & \textbf{38.3} & 36.5 & 33.2 & 36.0  \\
    XLSum (en) & 16.8 & 22.7 & 8.4 & \textbf{27.9} & 34.6 & {44.3} & \textbf{45.4} & 43.1 & 41.8 & 42.6  \\
    WikiLingua (en) & 6.5 & 6.4 & 5.9  &  \textbf{7.7} & \textbf{8.0} & 7.5 & 7.8 & 7.9 & 7.9 & 7.8  \\
    \midrule
     & \multicolumn{7}{c}{\textbf{Crosslingual Generation}} \\
    \midrule
    WikiLingua (es $\rightarrow$ en)  & 6.5 & 6.1 & 5.0 & \textbf{7.7} & 7.7 & 7.6 & 7.3 & 7.8 & 7.6 & \textbf{7.9}  \\
    WikiLingua (ru $\rightarrow$ en) & 10.2 & 17.5 & 0.7 & \textbf{18.9} & 29.9 & \textbf{35.7} & 25.1 & 27.9 & 31.7 & 30.8  \\
    WikiLingua (tr $\rightarrow$ en) & 10.1 & 20.0 & 7.7 & \textbf{21.2} & 31.1 & \textbf{38.8} & 31.5 & 26.8 & 36.7 & 28.2  \\
    WikiLingua (vi $\rightarrow$ en) & 7.7 & 14.5 & 2.2 & \textbf{16.2} & 28.9 & \textbf{32.9} & 22.9 & 22.7 & 31.0 & 28.5  \\
    \midrule 
    & & & \multicolumn{6}{c}{\textbf{Multilingual Generation}\ \textbf{\scriptsize{[SentencePiece-ROUGE-2]}}} \\
    \midrule
    MLSum (es) & 12.8 & \textbf{14.3} & 5.2 & 13.0 & 23.0 & 24.5 & 25.0 & 24.3 & \textbf{25.7} & 25.6  \\
    MLSum (de) & 13.6 & 21.3 & 3.9 & \textbf{22.6} & 35.2 & 41.4 & \textbf{44.1} & 43.5 & 43.3 & 43.7  \\
    XLSum (ar)  & 12.2 & \textbf{19.0} & 10.8 & 18.0 & 36.2 & 39.9 & 15.7 & 15.2 & \textbf{42.3} & 6.2  \\
    XLSum (bn) & 5.8 & 6.9 & 6.1 & \textbf{11.7} & 26.4 & 31.1 & 11.1 & 10.2 & \textbf{36.5} & 11.0  \\
    XLSum (ja) & 11.3 & 15.1 & 5.4  & \textbf{18.3} & 38.7 & 42.5 & 4.5  & 4.5 & \textbf{43.7} & 4.6  \\
    XLSum (id) & 16.8 & \textbf{20.4} & 9.0 & 20.1 & 35.5 & \textbf{43.5} & 41.1 & 41.6 & \textbf{43.5} & 40.8  \\
    XLSum (sw) & 16.7 & \textbf{24.5} & 11.5  & 15.4 & 32.7 & 36.4 & 37.0 & 37.4 & \textbf{40.7} & 36.3  \\
    XLSum (ko) & 16.1 & \textbf{18.2} & 7.9  & 17.6  & 33.8 & 37.3 & 20.3 & 19.5 & \textbf{45.0} & 19.9  \\
    XLSum (ru) & 12.6 & 16.1 & 10.8  & \textbf{19.1} & 30.3 & 38.3 & 18.1 & 17.8 & \textbf{38.6} & 17.7  \\
    XLSum (te) & 6.5 & 7.7 & 6.2  &\textbf{13.1} & 20.5 & 30.0 & 15.1 & 15.1 & \textbf{33.5} & 14.8  \\
    XLSum (th) & 6.7 & 8.6 & 5.2  & \textbf{13.3} & 23.4 & 29.5 & 13.5 & 13.7 & \textbf{34.3} & 13.1  \\
    XLSum (tr) & 15.2 & \textbf{17.7} & 8.0  & 16.8  & 33.3 & \textbf{42.4} & 30.3 & 30.4 & 42.3 & 29.7  \\
    XLSum (es) & 15.7 & \textbf{17.4} & 8.3  & 16.9 & 25.2 & \textbf{34.3} & 31.9 & 32.5 & 33.9 & 32.3  \\
    XLSum (vi) & 13.2 & 14.9 & 6.9  & \textbf{15.4} & 25.9 & \textbf{41.5} & 27.7 & 27.3 & 41.0 & 26.7  \\
    XLSum (hi) & 10.0 & 12.1 & 9.3  & \textbf{15.2} & 37.7 & \textbf{43.6} & 13.7 & 2.3 & 43.5 & 2.3  \\
    \midrule
    \end{tabular}
    \caption{\small{ROUGE-L and SentencePiece-ROUGE-L results on data-to-text and compression datasets. Best results in \textbf{bold}. Few-shot results lag behind finetuned results and the gap increases as tasks become more complex. The non-English performance mostly follows the trend that higher percentages of non-English pretraining data leads to better performance. Despite their much smaller size, encoder-decoder model frequently much larger decoder-only models after finetuning.}}
    \label{tab:results-table}
}
\end{table*}

\paragraph{Few-shot learning falls behind finetuning} For many generation tasks, including multilingual summarization tasks, we observe a large gap between finetuning and few-shot results, indicating that finetuning will play an important role when it comes to maximizing automatic scores. On data-to-text, the few-shot results follow a similar trend as in summarization, but the gap to the best finetuned results shrinks drastically. Moreover, the finetuning result do not always follow a trend according to scale or architecture. We hypothesize that multiple tasks have saturated to the metrics. If this is the case, approaching them as few-shot generation tasks may still yield insights but it is no longer productive to use them to benchmark finetuned models.

\paragraph{Finetuned decoder-only PLMs can match encoder-decoder performance with scale}
In summarization, finetuned decoder-only PLMs, such as \palmname-540B, closely match or exceeds the best reported prior results on all English generation tasks. This demonstrates that PLMs can make up their architectural disadvantage through its vastly increased scale.
While finetuning PLMs is computationally expensive, it serves as an important upper bound for few-shot predictions.

\paragraph{Multilingual generation capabilities are highly dependent on pretraining data} The PLMs evaluated are mostly pretrained on English corpora: 99+\% for T5, LongT5, \stmoename; 90\% for \palmname, \lamdaname; contrarily mT5 is explicitly pretrained in a multilingual corpus.\footnote{The language breakdown for GPT-3.5 is unknown.} \palmname achieves best results in 3 out of 4 English generation tasks which generate English text, even when the input is non-English. However, the much smaller mT5 bests the other models in 10 out of 14 non-English summarization tasks, and the relative difference between few-shot and finetuning is larger for non-English generation. This suggests that English-centric PLMs are better at processing non-English input than generating non-English output.

\paragraph{Analyzing the effects of input context length}
Tasks with long inputs suffer from models' limitation to process said inputs. Inputs are thus usually transformed (e.g. cropped, re-ranked, etc) to fit into the model. We found that a several of the evaluated tasks, such as WikiLingua and MLSum benefit from a longer input context in models even if the long-context model is smaller (i.e., LongT5 vs T5). In contrast, the performance is comparable for the rest of short-context tasks.

\begin{figure}[htb]
  \includegraphics[width=0.5\textwidth]{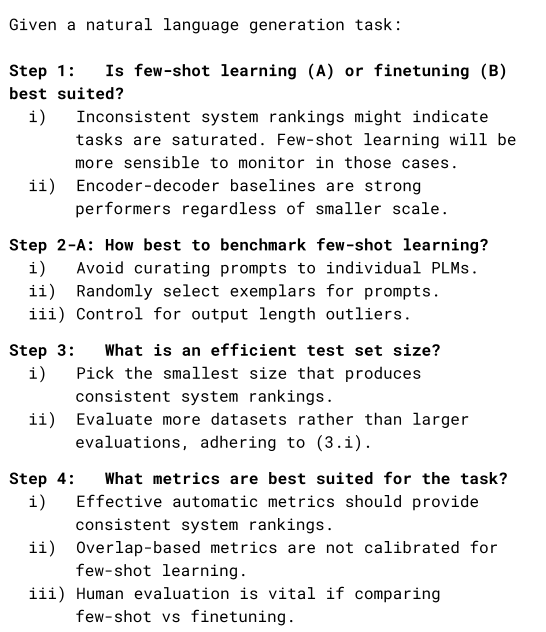}
  \caption{General recommendations when monitoring or benchmarking PLMs.}
  \label{fig:steps}
\end{figure}
\vspace{-5pt}
\section{Deriving Evaluation Practices}
\label{sec:practices}

Figure~\ref{fig:steps} summarizes the recommendations we developed from challenges we faced and our observed empirical results. These recommendations are best understood in the context of monitoring and benchmarking PLMs during training or inference.


\vspace{-5pt}

\paragraph{Comparable few-shot learning evaluation}
As mentioned in Section~\ref{sec:method}, our design choices were made to ensure that results are comparable across PLMs. Primarily, prompts were deliberately kept extremely simple and all few-shot exemplars were randomly sampled. While highly curated prompts or methods like chain-of-thought prompting can increase the performance considerably~\citep{Wei2022ChainOT}, it can also lead to overfitting to the particular model the prompt was developed on, in turn making a comparison to other models unfair and producing unrealistic expectations when people have single interactions with it.

\paragraph{Overlap-based metrics are not calibrated to evaluate few-shot learning} Few-shot generation suffers from not being able to predict output length properly given the few exemplars provided. While encoder-decoder models utilize end-of-string tokens, these are not always learned during decoder-only pretraining. To circumvent this issue, researchers rely on PLMs match to the few-shot format provided e.g. line-breaks that separate exemplars. We observed PLMs fail to follow the format a significant number of times, producing the largest allowed length on occasion. In our experiments, we tried to avoid very long outputs by trimming outputs to the 95-percentile length seen in the targets.\footnote{This simple method avoids discrepancies across PLMs which might have different maximum decoding lengths.}
Still, few-shot output lengths are on average 2-3 times the average target length while finetuned model's output average 80\% the average target length, across all tasks. Overlap metrics used in generation are sensitive to length \citep{sun-etal-2019-compare} making a natural disadvantage for few-shot learners. We do not recommend using overlap-based metrics to compare few-shot results without length normalization.


\paragraph{Computational costs can be decreased without sacrificing relative model performance}
The computational cost of evaluating large datasets, some with more than 10K examples, are prohibitive and perhaps unnecessary. To that end, we investigate 
if a model ranking can be produced, with a high degree of certainty, while only considering a random subset of the test set, saving compute cost to possibly evaluate on more tasks instead. 
To investigate this effect, we ran the following experiment: (1) Sample $n$ datapoints from a dataset and all corresponding model scores. (2) Following \citet{kocmi-etal-2021-ship} and \citet{graham-etal-2014-machine}, we perform Wilcoxon Rank Sum test~\citep{wilcoxon1946individual} to assess the stability of the ranking. (3) Repeat steps 1\&2 $k$ times and record the fraction of runs in which  models scores from any two models were not distinguishable from each other (those with a $p$-value of $> 0.05$). Since we are considering 10 model settings in this work, this experiment considers all 45 possible pairs.

\begin{figure}
  \includegraphics[width=0.48\textwidth]{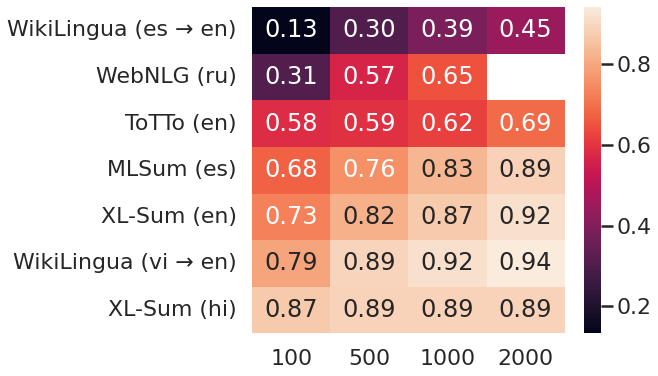}
  \caption{Empirical probability of p-value of Wilcoxon Rank Sum test $<$ 0.05 for any combination between 1-shot and finetuned models. 
  }
  \label{fig:wilcox-short}
\end{figure}

The result shown in Figure~\ref{fig:wilcox-short} provides insight into the required number of data points to produce rankings. For most datasets, we can produce stable model rankings with only 500 examples, some with as little as 100. Tasks where models achieve very similar scores tend to require more test examples, since smaller score differences require more examples to be distinguishable from each other~\citep{wei-jia-2021-statistical}.\footnote{Full results available in Appendix~\ref{sec:appendix}.}

\begin{figure}
  \includegraphics[width=0.48\textwidth]{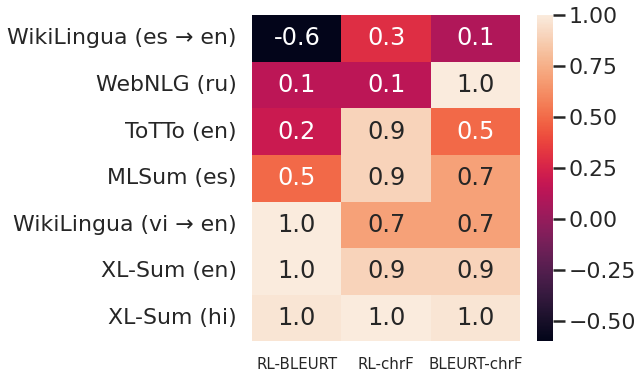}
  \caption{Spearman correlation coefficients between metrics: (SP)ROUGE-L, BLEURT and ChrF.}
  \label{fig:spearman}
\end{figure}

\paragraph{Analyzing metrics utility} We use different automated metrics to evaluate the generation quality of the models. These metrics attempt to capture the similarity between system generated output and the reference text. 
While ROUGE and chrF account for the lexical overlap, BLEURT is meant to compute the semantic similarity. It is important to understand the agreement between these metrics. We compute the the system-level agreement via Spearman correlation coefficient \cite{Spearman1987ThePA} between the scores given by the metrics to the fine-tuned set of models. Figure \ref{fig:spearman} shows the correlation between ROUGE-L (RL), BLEURT and ChrF. We observe that the metrics are highly correlated for most datasets. Similar to Figure~\ref{fig:wilcox-short}, on the tasks where the models have similar performance, we notice less correlation among the metrics. Such tasks are may have either saturated performance, e.g., ToTTo (en) or all models perform poorly, e.g., Wikilingua (es-> en). Due to the small differences between models, metrics fail to produce the same rankings.  

\section{Discussion and Reality Check}
\label{sec:discussion}
In line with our goal to provide a ``reality check'' via empirical and theoretical research, and to reflect on the ways in which reported performance improvements are meaningful, we want to situate our findings in the context of the broader NLP community. 
Openly accessible APIs and publicly available large models have led to increased attention on large pretrained models, but they have also led to a ``release-then-test'' philosophy where models are released without extensive evaluations. 
While the findings we present in this paper do not solve this issue, agreeing on a shared evaluation process could lead to more realistic claims about model performance (and shortcomings), and allow for a more accurate monitoring of models during training. 
\vspace{-5pt}
\paragraph{What claims can we not make?}

Empirical findings demonstrate that incorporating generation into NLU tasks via Chain-of-Thought leads to better model performance~\citep{Wei2022ChainOT,DBLP:journals/corr/abs-2210-09261}. 
Providing additional grounding via finetuning on instructions and aligning a model to human feedback leads to better task-specific performance without supervision~\citep{DBLP:conf/iclr/WeiBZGYLDDL22,DBLP:journals/corr/abs-2203-02155}. However, we lack the scientific methods to quantify these advances. While benchmarks provide an indication whether a model is performing better than a previous iteration, and projects like BIG-bench~\citep{DBLP:journals/corr/abs-2206-04615} and HELM~\citep{Liang2022HolisticEO} enable evaluation on a very wide range of possible tasks, they are also inherently limited. 

When benchmarking models in few-shot settings, especially models for which little information about their training data is available, it is hard to disambiguate model performance from memorization, i.e. if the examples were seen during pretraining. Instruction tuning further blur the line between finetuning and few-shot, which can lead to very different outputs and are not fully comparable. 
It is thus near impossible to make claims about \textit{why} a model is succeeding at one particular task without having access to its training data. 

As mentioned earlier, the target of this work is to derive best practices for comparing models in generation settings with constrained computational budgets, for example when monitoring a training model or when trying to compare on many different tasks. 
Our findings are grounded in much prior work that finds that metrics have a very high agreement with human judgments on the system-level~\citep[e.g.,][]{kocmi-etal-2021-ship}, but are essentially meaningless on the segment-level. For that reason, we cannot derive claims beyond these rankings about utility of a model or whether a particular model would actually produce useful outputs for a task. 
To derive such insights, we point to work on extrinsic evaluation which requires comprehensive human evaluations~\citep[e.g.,][]{DBLP:journals/corr/abs-2212-09746}.



\paragraph{How can our findings be applied to improve the status quo?}

Since the generation capabilities of PLMs are currently not extensively monitored or evaluated, we set out to derive best practices for how these evaluations can look. 
We found that many of the ``easy'' tasks, on which finetuned models saturate the metrics, still yield insights for few-shot approaches. 
We further identified the tension between doing a computationally expensive full evaluation on a dataset and adding more evaluation sets for different tasks or languages. 
Our findings suggest that evaluation on small subsets of more tasks can be beneficial to the overall results. 

To further motivate this suggestion, consider the following thought experiment: We have two tasks, A and B. At 500 examples, they have a risk of producing a ``wrong'' ranking of 10\%. At 1,000 examples, they have a risk of producing a wrong ranking of 5\%. These risks are not correlated, i.e., their covariance is 0. Given a computational budget of evaluating on 1,000 examples, the risk of only evaluating on one dataset is 5\%, and the risk of producing two wrong ratings after evaluating on A and B is only 1\%. While additional datasets introduce a larger risk of one individual dataset producing misleading results (18\% in this case), one can easily expand this argument to a whole portfolio of tasks to hedge against individual dataset risk~\citep{Stuart1959PortfolioSE}. Many existing NLU benchmarks like BIG bench~\citep{DBLP:journals/corr/abs-2206-04615} already follow such a strategy and we believe that generation evaluation, especially considering the additional risk due to metrics, should follow this approach for the use cases discussed in this work. 
To further minimize the individual dataset risk, they can be switched out once they saturate or their sample sizes increased. 

\section{Conclusion}
\label{sec:conclusion}
In this work, we produced an extensive evaluation of a diverse set of state-of-the-art pre-trained language models (PLMs) for 27 different multilingual generation tasks under few-shot learning and finetuning settings.
We discuss empirical results that help inform practitioners which tasks, methods and metrics are suitable. We  provide recommendations on how best to monitor conditional generation capabilities of PLMs, including how to fairly benchmark few-shot learning, automated metrics and their utility, and how to efficiently utilize computational resources.
We hope that such findings and recommendations could positively influence natural language evaluation in future work.

\section{Limitations}

In this work, we have presented results that help inform us what tasks, methods and metrics are best suited for monitoring as well as methodologies and empirical information about the current set of models. We provide detailed information of how these results can be reproduced, to the extend that research have access to the PLMs in question, but these results have limitations, in order to reduce costs, many languages were not evaluated which might have left unforeseen patterns not discussed in this work. Moreover, few-shot learning, in particular, could exhibit large variance if different prompts were chosen, or a different set of exemplars chosen. Because of the high costs involved our work does not explore the performance difference when multiple sets of hyper-parameters were chosen. 

On the conceptual level, we make the assumption that system-level improvements on our tasks translate to downstream usefulness. While prior work suggests that this is the case, tools like chatGPT have significantly expanded the possible application space beyond the realm of ``typical'' NLP tasks, and we don't know how well our findings generalize to this space of tasks.

\section{Ethics Statement}

This paper focuses on conditional generation tasks where models are free to generate long text sequences. Typical issues associated with text generation such as hallucinations, memorization of private information publicly available, toxic and discriminatory language, or sensitive generated content could and are likely to arise. 
measuring the extent to which these issues occur is a necessary and crucial additional dimension of model evaluation which we do not include in this work, which should be seen as supplemental. 

\bibliography{anthology,custom}

\clearpage
\appendix
\section{Additional empirical results}

Table~\ref{tab:rouge2-table}, 
Table~\ref{tab:chrf-table} and
Table~\ref{tab:bleurt-table} report ROUGE-2 and
BLEURT and ChrF results respectively for all tasks. These results are in line with the discussed results in  \ref{sec:results}

\begin{figure*}
  \includegraphics[width=\textwidth]{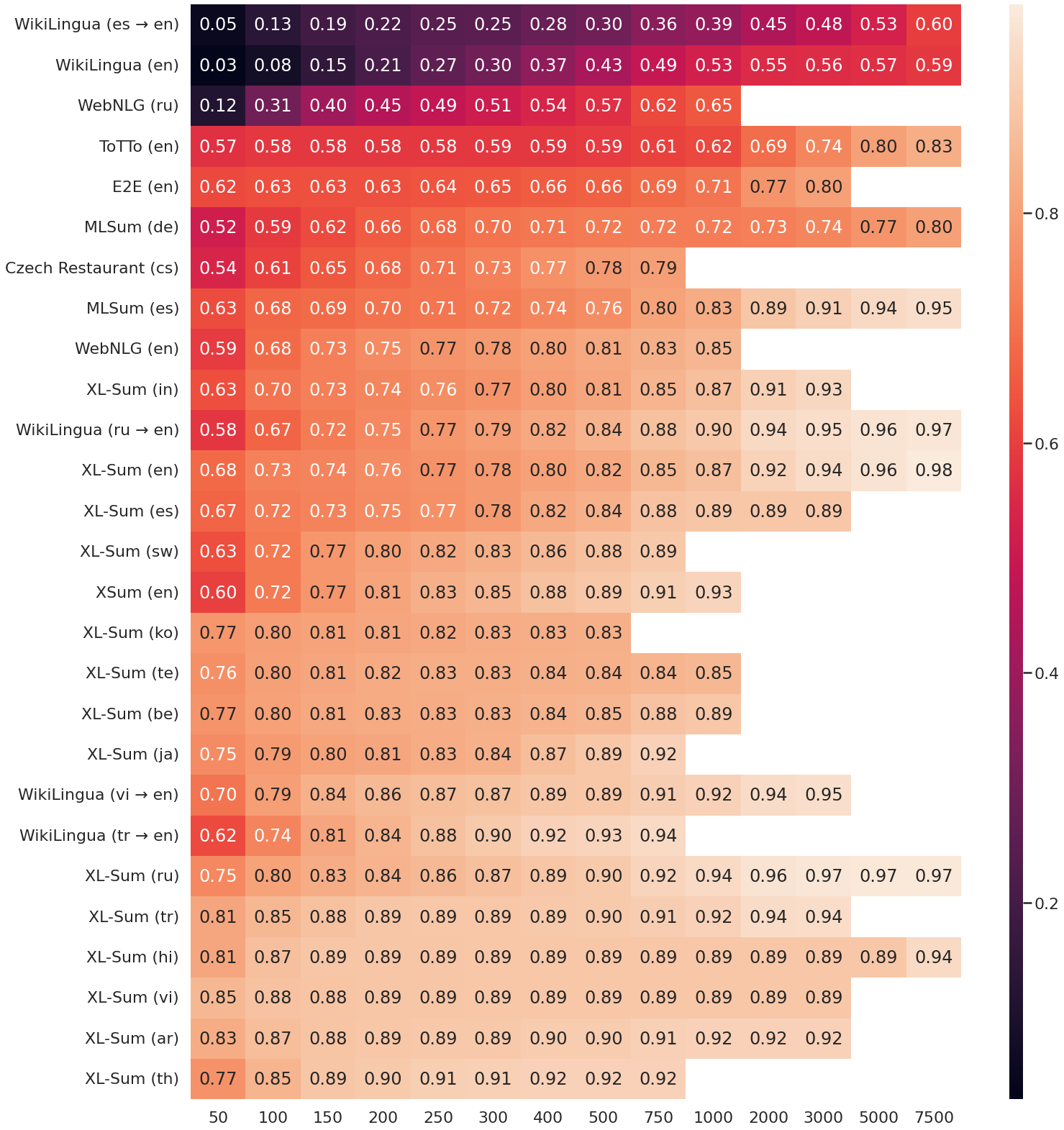}
  \caption{Empirical probability of p-value of Wilcoxon Rank Sum test $<$ 0.05 for any combination between 1-shot and finetuned models.}
\end{figure*}
\label{sec:appendix}

\begin{table*}[h]
\footnotesize{
    \centering
    \begin{tabular}{p{3cm}cccc|cccccc}
    \toprule
    & \multicolumn{4}{c}{One shot} & \multicolumn{6}{c}{Finetuning} \\
    \cmidrule(l{3pt}r{3pt}){2-5} \cmidrule(l{3pt}r{3pt}){6-11}
    Task & \scriptsize{\makecell[c]{\palmname \\8B}} & \scriptsize{\makecell[c]{\palmname \\540B}} & \scriptsize{\makecell[c]{\lamdaname \\137B}} & \scriptsize{\makecell[c]{GPT-3.5 \\175B}}  & \scriptsize{\makecell[c]{\palmname \\8B}} & \scriptsize{\makecell[c]{\palmname \\540B}} & \scriptsize{\makecell[c]{\stmoename \\32B}} &  \scriptsize{\makecell[c]{T5 \\11B}} & \scriptsize{\makecell[c]{mT5 \\11B}} & \scriptsize{\makecell[c]{LongT5 \\3B}}   \\
    \midrule
     & \multicolumn{7}{c}{\textbf{Data-To-Text}} \\
    \midrule
    E2E (en) & 26.7 & 37.3 & 4.2 & 37.9 & 45.7 & 45.3 & 44.2 & 45.2 & 45.5 & 46.3  \\
    WebNLG (en) & 33.8 & 45.8 & 5.4 & 46.0 & 47.7 & 49.2 & 47.6 & 39.6 & 35.8 & 48.8  \\
    ToTTo (en) & 26.4 & 37.8 & 3.2 & 38.1 & 53.9 & 55.9 & 55.2 & 54.1 & 53.3 & 54.5  \\
    Czech Restaurant (cs) & 7.9 & 18.1 & 0.9 & 22.3 & 30.2 & 30.6 & 25.4 & 28.8 & 25.0 & 29.9  \\
    WebNLG (ru) & 4.9 & 16.5 & 1.2 & 16.8 & 22.4 & 23.4 & 13.0 & 23.1 & 23.2 & 24.2  \\
    \midrule
     & \multicolumn{7}{c}{\textbf{English Generation}} \\
    \midrule
    XSum (en) & 8.0 & 14.4 & 3.4 & 19.9 & 16.3 & 21.2 & 22.8 & 21.0 & 17.5 & 20.8  \\
    XLSum (en) & 6.5 & 11.7 & 2.7 & 17.0 & 19.6 & 29.5 & 30.6 & 28.0 & 26.5 & 27.4  \\
    WikiLingua (en) & 0.7 & 0.4 & 0.7 & 0.9 & 0.4 & 0.4 & 0.4 & 0.4 & 0.4 & 0.4  \\
    \midrule
     & \multicolumn{7}{c}{\textbf{Crosslingual Generation}} \\
    \midrule
    WikiLingua (es $\rightarrow$ en) & 0.7 & 0.5 & 0.4 & 0.6 & 0.4 & 0.4 & 0.3 & 0.4 & 0.4 & 0.4  \\
    WikiLingua (ru $\rightarrow$ en) & 3.1 & 6.8 & 0.1 & 7.8 & 14.0 & 18.7 & 12.0 & 12.7 & 15.1 & 14.4  \\
    WikiLingua (tr $\rightarrow$ en) & 3.1 & 8.7 & 2.1 & 10.1 & 16.6 & 23.0 & 17.7 & 13.8 & 20.8 & 14.3  \\
    WikiLingua (vi $\rightarrow$ en) & 2.4 & 5.5 & 0.4 & 6.8 & 13.4 & 0.4 & 10.3 & 9.7 & 14.8 & 13.2  \\
    \midrule 
    & & & \multicolumn{6}{c}{\textbf{Multilingual Generation}\ \textbf{\scriptsize{[SentencePiece-ROUGE-2]}}} \\
    \midrule
    MLSum (es) & 3.7 & 4.5 & 0.7 & 4.9 & 10.7 & 0.7 & 13.1 & 12.1 & 13.5 & 13.6  \\
    MLSum (de) & 8.8 & 16.8 & 1.2 & 16.9 & 26.9 & 33.4 & 36.5 & 36.1 & 35.9 & 36.3  \\
    XLSum (ar) & 4.5 & 11.7 & 2.4 & 9.6 & 25.8 & 30.0 & 1.9 & 2.0 & 32.1 & 0.6  \\
    XLSum (bn) & 1.0 & 1.8 & 0.5 & 2.7 & 18.5 & 23.5 & 0.2 & 0.1 & 29.4 & 0.1  \\
    XLSum (ja) & 4.0 & 6.7 & 0.3 & 9.6 & 27.1 & 31.8 & 0.1 & 0.1 & 31.5 & 0.1  \\
    XLSum (id) & 7.2 & 11.3 & 3.7 & 12.6 & 25.0 & 33.0 & 30.5 & 30.7 & 32.7 & 30.3  \\
    XLSum (sw) & 7.9 & 16.2 & 6.5 & 6.6 & 22.7 & 26.8 & 27.8 & 27.6 & 31.3 & 26.8  \\
    XLSum (ko) & 6.9 & 9.6 & 1.6 & 9.4 & 23.0 & 26.7 & 4.1 & 3.7 & 34.9 & 4.0  \\
    XLSum (ru) & 6.0 & 9.2 & 3.7 & 10.9 & 20.8 & 29.4 & 4.5 & 6.1 & 29.6 & 6.1  \\
    XLSum (te) & 2.4 & 3.3 & 1.1 & 4.8 & 13.3 & 22.7 & 3.2 & 3.2 & 26.5 & 3.3  \\
    XLSum (th) & 2.9 & 4.0 & 0.3 & 6.2 & 16.4 & 22.5 & 2.4 & 2.5 & 26.9 & 2.4  \\
    XLSum (tr) & 7.5 & 10.5 & 3.2 & 10.7 & 23.7 & 32.7 & 17.6 & 17.8 & 32.2 & 17.7  \\
    XLSum (es) & 5.8 & 9.0 & 3.1 & 9.6 & 14.2 & 23.7 & 20.1 & 20.6 & 23.0 & 20.6  \\
    XLSum (vi) & 4.8 & 6.8 & 1.5 & 7.5 & 20.2 & 35.9 & 11.9 & 13.2 & 35.5 & 13.1  \\
    XLSum (hi) & 4.4 & 6.4 & 1.8 & 7.0 & 29.0 & 35.7 & 1.8 & 0.0 & 35.4 & 0.0  \\
    \midrule
    \end{tabular}
    \caption{ROUGE-2 and SentencePiece-ROUGE-2 results in data-to-text, English and multilingual generation datasets. }
    \label{tab:rouge2-table}
}
\end{table*}

\begin{table*}[h]
\footnotesize{
    \centering
    \begin{tabular}{p{3cm}cccc|cccccc}
    \toprule
    & \multicolumn{4}{c}{One shot} & \multicolumn{6}{c}{Finetuning} \\
    \cmidrule(l{3pt}r{3pt}){2-5} \cmidrule(l{3pt}r{3pt}){6-11}
    Task & \scriptsize{\makecell[c]{\palmname \\8B}} & \scriptsize{\makecell[c]{\palmname \\540B}} & \scriptsize{\makecell[c]{\lamdaname \\137B}} & \scriptsize{\makecell[c]{GPT-3.5 \\175B}}  & \scriptsize{\makecell[c]{\palmname \\8B}} & \scriptsize{\makecell[c]{\palmname \\540B}} & \scriptsize{\makecell[c]{\stmoename \\32B}} &  \scriptsize{\makecell[c]{T5 \\11B}} & \scriptsize{\makecell[c]{mT5 \\11B}} & \scriptsize{\makecell[c]{LongT5 \\3B}}   \\
    \midrule
     & \multicolumn{7}{c}{\textbf{Data-To-Text}} \\
    \midrule
    E2E (en) & 46.1 & 57.8 & 15.1 & 57.8 & 62.5 & 61.9 & 61.1 & 62.1 & 60.9 & 61.8  \\
    WebNLG (en) & 47.5 & 61.8 & 17.0 & 61.8 & 63.6 & 65.2 & 64.1 & 59.4 & 55.8 & 65.4  \\
    ToTTo (en) & 43.5 & 55.8 & 12.6 & 55.2 & 67.5 & 69.4 & 68.3 & 67.3 & 66.8 & 67.7  \\
    Czech Restaurant (cs) & 17.6 & 35.6 & 7.9 & 41.5 & 48.0 & 48.1 & 36.6 & 38.2 & 44.1 & 40.1  \\
    WebNLG (ru) & 21.8 & 45.5 & 15.3 & 45.3 & 62.7 & 62.6 & 24.4 & 31.8 & 63.5 & 32.1  \\
    \midrule
     & \multicolumn{7}{c}{\textbf{English Generation}} \\
    \midrule
    XSum (en) & 24.6 & 32.7 & 18.5 & 37.6 & 34.4 & 38.9 & 41.3 & 39.3 & 36.8 & 39.0  \\
    XLSum (en) & 23.8 & 29.9 & 13.9 & 33.5 & 35.8 & 45.8 & 47.0 & 46.0 & 43.9 & 44.4  \\
    WikiLingua (en) & 14.7 & 14.3 & 15.4 & 15.8 & 15.1 & 14.6 & 15.2 & 15.8 & 15.8 & 15.3  \\
    \midrule
     & \multicolumn{7}{c}{\textbf{Crosslingual Generation}} \\
    \midrule
    WikiLingua (es $\rightarrow$ en) & 16.8 & 13.2 & 10.2 & 14.6 & 14.2 & 14.9 & 13.0 & 15.1 & 14.8 & 15.7  \\
    WikiLingua (ru $\rightarrow$ en) & 19.1 & 21.5 & 1.5 & 22.5 & 30.6 & 35.9 & 24.0 & 28.8 & 34.3 & 33.3  \\
    WikiLingua (tr $\rightarrow$ en) & 19.3 & 24.6 & 12.3 & 26.7 & 34.4 & 39.4 & 32.6 & 30.8 & 39.0 & 32.7  \\
    WikiLingua (vi $\rightarrow$ en) & 16.4 & 19.9 & 4.4 & 21.3 & 31.8 & 14.2 & 23.2 & 27.6 & 32.3 & 32.5  \\
    \midrule 
    & \multicolumn{7}{c}{\textbf{Multilingual Generation} } \\
    \midrule
    MLSum (es) & 21.3 & 22.9 & 5.3 & 20.6 & 28.0 & 18.4 & 29.1 & 28.7 & 30.7 & 30.3  \\
    MLSum (de) & 28.9 & 37.0 & 5.7 & 34.4 & 41.9 & 48.8 & 50.9 & 50.5 & 50.1 & 51.5  \\
    XLSum (ar) & 14.2 & 22.7 & 11.4 & 24.4 & 35.0 & 39.6 & 0.2 & 0.2 & 41.6 & 0.1  \\
    XLSum (bn) & 10.3 & 12.7 & 4.5 & 17.5 & 28.6 & 34.0 & 0.0 & 0.0 & 37.8 & 0.0  \\
    XLSum (ja) & 8.8 & 11.6 & 1.2 & 13.8 & 26.0 & 31.3 & 0.8 & 0.9 & 30.6 & 0.9  \\
    XLSum (id) & 21.0 & 26.0 & 9.0 & 26.2 & 36.8 & 45.3 & 43.2 & 42.3 & 43.0 & 43.4  \\
    XLSum (sw) & 24.0 & 33.0 & 15.0 & 21.5 & 36.2 & 42.0 & 40.1 & 41.1 & 44.4 & 41.6  \\
    XLSum (ko) & 6.9 & 9.4 & 1.6 & 10.0 & 18.0 & 21.8 & 1.4 & 1.2 & 27.5 & 1.4  \\
    XLSum (ru) & 15.0 & 19.8 & 9.4 & 26.5 & 29.1 & 38.6 & 14.4 & 20.1 & 40.3 & 19.9  \\
    XLSum (te) & 11.3 & 13.6 & 4.7 & 16.8 & 18.0 & 29.8 & 0.3 & 0.2 & 30.4 & 0.3  \\
    XLSum (th) & 14.7 & 16.8 & 4.4 & 21.5 & 27.1 & 33.4 & 0.3 & 0.3 & 33.9 & 0.3  \\
    XLSum (tr) & 20.3 & 24.9 & 6.2 & 24.5 & 32.7 & 43.1 & 31.2 & 33.1 & 42.6 & 33.8  \\
    XLSum (es) & 19.0 & 22.9 & 7.3 & 22.0 & 24.5 & 33.4 & 31.5 & 31.9 & 32.6 & 32.8  \\
    XLSum (vi) & 10.9 & 13.1 & 2.4 & 14.2 & 21.8 & 37.1 & 16.9 & 20.2 & 34.3 & 21.1  \\
    XLSum (hi) & 12.2 & 15.1 & 6.6 & 18.8 & 33.2 & 39.6 & 0.2 & 0.0 & 39.1 & 0.0  \\
    \midrule
    \end{tabular}
    \caption{ChrF results in data-to-text, English and multilingual generation datasets.}
    \label{tab:chrf-table}
}
\end{table*}

\begin{table*}[h]
\footnotesize{
    \centering
    \begin{tabular}{p{3cm}cccc|cccccc}
    \toprule
    & \multicolumn{4}{c}{One shot} & \multicolumn{6}{c}{Finetuning} \\
    \cmidrule(l{3pt}r{3pt}){2-5} \cmidrule(l{3pt}r{3pt}){6-11}
    Task & \scriptsize{\makecell[c]{\palmname \\8B}} & \scriptsize{\makecell[c]{\palmname \\540B}} & \scriptsize{\makecell[c]{\lamdaname \\137B}} & \scriptsize{\makecell[c]{GPT-3.5 \\175B}}  & \scriptsize{\makecell[c]{\palmname \\8B}} & \scriptsize{\makecell[c]{\palmname \\540B}} & \scriptsize{\makecell[c]{\stmoename \\32B}} &  \scriptsize{\makecell[c]{T5 \\11B}} & \scriptsize{\makecell[c]{mT5 \\11B}} & \scriptsize{\makecell[c]{LongT5 \\3B}}   \\
    \midrule
     & \multicolumn{7}{c}{\textbf{Data-To-Text}} \\
    \midrule
    E2E (en) & 60.0 & 71.8 & 44.2 & 72.3 & 76.5 & 75.8 & 75.1 & 76.4 & 75.9 & 76.2  \\
    WebNLG (en) & 62.3 & 74.5 & 43.5 & 74.7 & 75.8 & 76.8 & 75.6 & 71.2 & 67.8 & 76.3  \\
    ToTTo (en) & 56.9 & 69.4 & 33.1 & 69.5 & 76.8 & 77.9 & 77.0 & 76.6 & 76.8 & 76.7  \\
    Czech Restaurant (cs) & 34.7 & 66.8 & 32.2 & 72.2 & 75.8 & 74.4 & 48.8 & 51.8 & 72.9 & 48.8  \\
    WebNLG (ru) & 39.2 & 67.8 & 19.7 & 66.9 & 77.5 & 78.0 & 25.9 & 29.8 & 78.4 & 29.9  \\
    \midrule
     & \multicolumn{7}{c}{\textbf{English Generation}} \\
    \midrule
    XSum (en) & 43.0 & 46.9 & 28.2 & 53.4 & 51.0 & 55.5 & 58.5 & 56.4 & 53.2 & 56.4  \\
    XLSum (en) & 32.7 & 41.1 & 22.0 & 51.1 & 52.6 & 61.9 & 63.0 & 61.8 & 60.3 & 60.8  \\
    WikiLingua (en) & 33.3 & 34.0 & 27.9 & 34.3 & 32.2 & 32.2 & 31.3 & 32.4 & 32.6 & 32.1  \\
    \midrule
     & \multicolumn{7}{c}{\textbf{Crosslingual Generation}} \\
    \midrule
    WikiLingua (es $\rightarrow$ en) & 32.9 & 33.7 & 16.8 & 33.4 & 32.3 & 32.6 & 31.0 & 32.5 & 33.0 & 32.5  \\
    WikiLingua (ru $\rightarrow$ en) & 38.8 & 43.3 & 6.4 & 45.9 & 50.6 & 54.4 & 46.4 & 49.1 & 52.3 & 51.5  \\
    WikiLingua (tr $\rightarrow$ en) & 39.3 & 44.0 & 19.3 & 46.4 & 49.2 & 54.4 & 48.6 & 45.4 & 52.8 & 46.8  \\
    WikiLingua (vi $\rightarrow$ en) & 35.6 & 38.2 & 5.7 & 40.8 & 50.6 & 32.8 & 45.4 & 45.5 & 51.4 & 50.0  \\
    \midrule 
    & \multicolumn{7}{c}{\textbf{Multilingual Generation} } \\
    \midrule
    MLSum (es) & 21.0 & 21.5 & -1.3 & 26.7 & 30.6 & 6.3 & 25.9 & 24.2 & 32.0 & 27.0  \\
    MLSum (de) & 39.4 & 50.1 & 4.6 & 49.3 & 57.8 & 63.4 & 62.5 & 61.6 & 61.5 & 61.4  \\
    XLSum (ar) & 19.8 & 28.5 & 2.5 & 27.7 & 44.6 & 50.1 & 2.8 & 3.5 & 53.0 & 4.3  \\
    XLSum (bn) & 31.8 & 41.8 & 0.2 & 27.4 & 46.6 & 57.1 & 2.9 & 3.6 & 62.7 & 2.8  \\
    XLSum (ja) & 28.1 & 31.4 & -1.2 & 34.7 & 47.0 & 52.2 & -0.3 & -0.3 & 53.0 & -0.3  \\
    XLSum (id) & 41.2 & 47.4 & 9.5 & 53.8 & 58.7 & 68.0 & 61.4 & 65.5 & 66.8 & 65.6  \\
    XLSum (sw) & 25.5 & 36.3 & 14.3 & 24.0 & 45.8 & 52.9 & 48.6 & 50.2 & 59.1 & 48.9  \\
    XLSum (ko) & 25.6 & 31.6 & -0.3 & 33.0 & 40.5 & 47.1 & 0.8 & 1.6 & 54.6 & 1.4  \\
    XLSum (ru) & 30.1 & 37.3 & 3.2 & 33.0 & 47.9 & 59.6 & 14.2 & 16.7 & 58.0 & 17.0  \\
    XLSum (te) & 29.6 & 35.0 & 6.5 & 22.7 & 32.0 & 49.1 & 10.9 & 11.5 & 51.6 & 12.4  \\
    XLSum (th) & 22.0 & 27.2 & -0.1 & 16.3 & 31.9 & 43.6 & -1.1 & -0.9 & 46.0 & -1.1  \\
    XLSum (tr) & 30.8 & 34.5 & 3.3 & 40.0 & 49.8 & 63.8 & 21.4 & 26.4 & 62.5 & 26.3  \\
    XLSum (es) & 21.2 & 26.3 & 0.0 & 30.6 & 31.5 & 46.2 & 33.3 & 36.1 & 45.2 & 35.7  \\
    XLSum (vi) & 14.5 & 14.5 & -1.6 & 16.4 & 24.7 & 46.5 & -4.0 & -4.6 & 45.0 & -4.5  \\
    XLSum (hi) & 33.9 & 40.4 & 7.0 & 33.7 & 50.7 & 57.5 & 5.7 & 4.6 & 57.3 & 4.6  \\
    \midrule
    \end{tabular}
    \caption{BLEURT results in data-to-text, English and multilingual generation datasets.}
    \label{tab:bleurt-table}
}
\end{table*}

\section{Technical details}
\label{sec:appendix_technical}
Finetuning and inference was done in the \textbf{t5x} framework for public and closed access models. Few-shot learning task methodology is well described in \ref{sec:method}, for public access models inference was done via their respective public API, whilst all other models were loaded from the standard checkpoint into TPU accelerators and inference was done on batches of 64. Finetuning was carried out in TPU accelerators, for \palmname we used a constant learning rate of $5{\times}10^{-5}$, 20x smaller than during pretraining and reset the optimizer's~(Adafactor) accumulators, for T5, mT5 and LongT5 we used a constant learning rate of $1{\times}10^{-4}$.

\section{Computational Cost and Environmental Impact}
In our work we benchmark 27 generation tasks which require a substantial amount of computational resources. Inference on PLMs is exceedingly more efficient than finetuning. We report the number of test examples across all tasks to be 194,844. Inference over this number of examples times 10 models evaluated amounts to 2 million inference examples. Finetuning on the other hand, requires all parameters to be trained and training dataset sizes are considerably larger. We estimate the compute usage for finetuning to be 256 TPU v3 chips for 200 hours. One of the goals of this work is to encourage benchmarking in the future to avoid these costs by more efficiently selecting smaller test size and persuade researchers to only evaluate finetuning approaches when suitable.

\end{document}